\pdfoutput=1

\documentclass[11pt]{article}

\usepackage[]{acl}

\usepackage{times}
\usepackage{latexsym}

\usepackage[T1]{fontenc}

\usepackage[utf8]{inputenc}

\usepackage{microtype}

\usepackage{amsfonts}       
\usepackage{nicefrac}       
\usepackage{amsmath}

\usepackage{pgfplots}
\pgfplotsset{compat=1.17}
\usepackage{verbatim}

\usepackage{booktabs}

\usepackage{appendix}

\title{Distributionally Robust Recurrent Decoders\\ with Random Network Distillation}

\author{Antonio Valerio Miceli Barone \\
  University of Edinburgh \\
  \texttt{amiceli@ed.ac.uk} \\\And
  Alexandra Birch \\
  University of Edinburgh \\
  \texttt{a.birch@ed.ac.uk} \\\And
  Rico Sennrich \\
  Universität Zürich \\
  \texttt{sennrich@cl.uzh.ch} \\
  }

\begin{document}
\maketitle
\begin{abstract}
Neural machine learning models can successfully model language that is similar to their training distribution, but they are highly susceptible to degradation under distribution shift, which occurs in many practical applications 
when processing out-of-domain (OOD) text
.
This has been attributed to "shortcut learning": relying on weak correlations over arbitrary large contexts.
We propose a method based on OOD detection with Random Network Distillation to allow an autoregressive language model to automatically disregard OOD context during inference, smoothly transitioning towards a less expressive but more robust model as the data becomes more OOD, while retaining its full context capability when operating in-distribution.
We apply our method to a GRU architecture, demonstrating improvements on multiple language modeling (LM) datasets.
\end{abstract}

\section{Introduction}
\label{SEC:INTRO}

Neural language models have become the main component of modern natural language processing systems, with larger and larger models being used as feature extractors for downstream tasks \citep{devlin-etal-2019-bert}, as probability estimators for ranking and ensembling \citep{Gulcehre2015fusion} or as language generators \citep{DBLP:journals/corr/BahdanauCB14, vaswani2017transformer, brown2020language}.

Despite their success, neural machine learning models can suffer large performance degradation when they are applied to out-of-domain data which is substantially different than their training data \citep{Lapuschkin2019, Hupkes2019, Recht2019}.


Unlike the older statistical language models, Recurrent LMs (RNNLMs) \citep{mikolov2010recurrent} and their successors Transformers LMs \citep{vaswani2017transformer} can consider the entire prefix of a sentence when predicting or generating the next token.
By being able to relate a very high-dimensional input to the output, these models can learn many subtle correlations which are highly useful as long as the input is in-distribution, unfortunately these correlations tend to be brittle to distribution shift, causing a model that depends on them to go astray.
This phenomenon is known as "shortcut learning" \citep{geirhos2020shortcut} and it has been found to also occur in humans and animals, but it is especially prevalent in artificial neural networks.
Research on this problem has explored models  invariant or equivariant w.r.t. certain transformations by means of compositional representations \citep{Sabour2017, Soulos2019, liu2020compositional}, causal modeling \citep{scholkopf2021toward}, or both \citep{arjovsky2019invariant, Krueger2020}, but these works focus on classification tasks often on synthetic datasets and can't be straightforwardly applied to black-box language models.
Approaches specific to LMs have focused on robustness where the data domains are known and represented in the training data \citep{oren-etal-2019-distributionally, Gerstenberger2020}.


In this work we propose a method that uses Random Network Distillation (RND) \citep{burda2018exploration} to dynamically adapt the amount of context that the model relies upon during inference based on an estimate of how much this context is out-of-distribution (OOD).
This way the model can still make use of all available context when operating within a familiar context space, exploiting long-distance weak correlations, but it reduces to a less expressive and more robust model when operating OOD, relying only on the strongest correlations.
As a proof of concept we implement our approach on a GRU recurrent language model \citep{cho-etal-2014-learning}. 
While Transformer decoders outperform RNNs when trained on large training sets, RNNs remain competitive on smaller datasets ($< 10^7$ tokens) where OOD phenomena are easier to measure, furthermore they are easier to optimize, simplifying architecture and hyperparameter search.
We evaluate our method on language modeling 
tasks on English datasets, obtaining improvements when evaluating on eight OOD domains.
We report additional preliminary sequence-to-sequence results on Transformer-RNN models \citep{zhang-etal-2018-accelerating} in appendix \ref{SEC:SEQ2SEQ}. We leave extensions of our method to full Transformers as future research.

\section{Background}
\label{SEC:BACKGROUND}

\paragraph{Recurrent Language Model}
\label{SEC:BACKGROUND:RNNLM}

Given a sequence $x(t)$ of tokens encoded as one-hot vectors, an  autoregressive causal recurrent language model estimates at each step $t$ a probability distribution $\Pr(x(t+1) | x(0), \dots, x(t)) = y(t+1)$ over the next token conditional on the observed prefix which is summarized as a fixed-dimensional state $h(t+1) \in \mathcal{R}^d$ computed according to the recurrence relation:

\small
\begin{align}
    u(t) &= \text{Emb}(x(t), \theta) \\
    h(0) &= 0^{\otimes d} \\
    h(t+1) &= \text{RNN}(h(t), u(t), \theta) \label{EQ:RNN:STATE} \\
    y(t+1) &= \text{Proj}(h(t+1), \theta)
\end{align}
\normalsize
where $\text{Emb}$ is an embedding layer, $\text{RNN}$ is a recurrent cell (in our case, a GRU), $\text{Proj}$ is a readout layer (we use a mixture-of-softmaxes layer \citep{yang2018breaking}) and $\theta$ represents all the trainable parameters.
The initial state $h(0)$ is fixed at zero.

\begin{figure}
    \centering
    \includegraphics[scale=0.55]{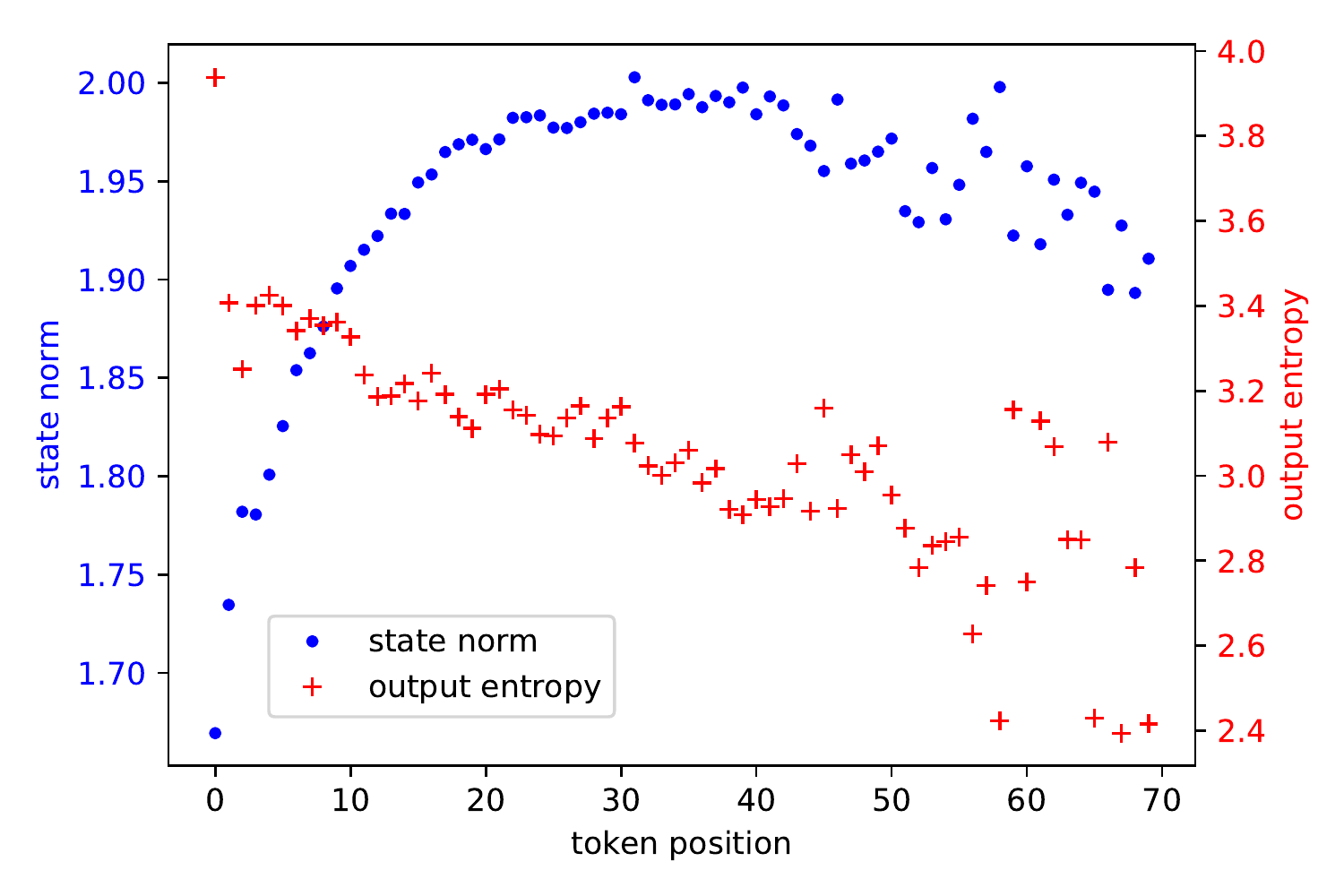}
    \caption{L2-norm of top GRU state (blue dots) and output softmax entropy (red crosses) over BPE token position, averaged over in-domain test set (Penn Treebank). Sentences are evaluated independently starting from the zero state. Norm and entropy correlate at $-0.58$.
    Model trained on Penn Treebank (sec. \ref{SEC:EXPERIMENTS:PERPLEXITY}).
    }
    \label{fig:entropy_and_norm_over_position}
\end{figure}

An interesting property of this model is that close to the beginning of the sequence the state vector $h(t)$ has a small norm, and the entropy of the predicted token distribution is usually high because many tokens are plausible, while as more and more tokens are observed the state norm grows (token embeddings are approximately "added" to the state \citep{levy2018lstm}) up to a point, and at the same time the entropy of the predicted token distribution decreases as the model becomes more confident of its prediction due to the larger observed context (Figure \ref{fig:entropy_and_norm_over_position}).
Indeed, in a softmax readout layer:

\small
\begin{align*}
    \text{Proj}(h) = \text{softmax}(W \cdot h + b)    
\end{align*}
\normalsize
where $W$ is the output projection matrix and $b$ is the output bias vector, 
increasing the norm of the state vector $h$ will usually cause the probability distribution to become sharper unless $W \cdot h$ happens to approximately cancel out the bias vector
$b$, which in high dimensions requires a rather specific alignment.
A mixture-of-softmaxes readout also exhibits this property.
Furthermore, it has been observed that the state of a RNN is usually dominated by the most recently observed inputs as the contribution of past inputs decreases exponentially over time \citep{jaeger2001echo, pascanu2013difficulty, levy2018lstm, zhang-sennrich-2019-lightweight}.
Therefore, we hypothesize that the norm of the state vector corresponds to the amount of context that the model is considering for its future predictions, 
and this in turn controls the confidence of the model in its predictions.

\paragraph{Random Network Distillation}
\label{SEC:BACKGROUND:RND}
In order to estimate how much the state of our RNNLM has deviated from the training distribution we choose the Random Network Distillation (RND) approach \citep{burda2018exploration, Ciosek2020Conservative}.
Given a representation $h$, we define an OOD detector as

\small
\begin{equation}
    \text{OOD}(h) = |T(h) - S(h, \phi)|^2
    \label{EQ:RND:OOD}
\end{equation}
\normalsize

where $T(h)$ is a randomly initialized and frozen feed-forward teacher network that pseudo-randomly maps the state $h$ to a high-dimensional output and $S(h, \phi)$ is a feed-forward student network with parameters $\phi$ trained to copy the teacher by minimizing eq. \ref{EQ:RND:OOD} on the training set.
At inference time the distillation error of eq. \ref{EQ:RND:OOD} provides an OOD estimate of $h$.
This works by deliberately exploiting the fragility of neural networks w.r.t. distribution shift: while in principle the student could learn to copy the teacher for all possible inputs, in practice it only learns to do so on the training set (in-domain by definition) and becomes increasingly uncorrelated to it as the input becomes more OOD.
See \citet{Ciosek2020Conservative} for an extensive analysis.
We chose this method because it can be applied to internal representations, is completely unsupervised and does not require any OOD tuning data.
RND has been proposed initially in the context of reinforcement learning where the OOD signal can be used as a ``curiosity'' reward to stimulate exploration, and it has been subsequently studied in the context of OOD estimation for image classification.
To our knowledge, we are the first to apply it to NLP, and to use it to actively compensate for distribution shift rather than just measure it.

\section{Proposed approach}
\label{SEC:APPROACH}
Our approach consists of estimating how much out-of-distribution the state of the model is and scaling it towards the all-zero initial state accordingly, effectively purging the OOD context out of the memory of the model and forcing it to rely only on the strongest, usually short-distance, correlations that survive the purge.
As the state is pushed towards zero, the model also becomes more conservative in its predictions, avoiding the typical overconfidence of neural networks in OOD conditions.
Specifically, for our language modeling experiments, we train a GRU RNNLM as usual, then we freeze it and train a RND OOD estimator on the RNNLM states on the same training set. Then during inference we modify the recurrence relation (eq. \ref{EQ:RNN:STATE}) to

\small
\begin{align}
    \tilde{h} &= \text{RNN}(h(t), u(t), \theta) \\
    h(t+1) &= \tilde{h} \cdot \alpha \exp(- \beta \cdot \text{OOD}(\tilde{h})) 
\end{align}
\normalsize
where we use a simple exponential scaling with $\alpha$ and $\beta$ hyperparameters\footnote{$\alpha$ can also be tuned by SGD on the training set, but we found this to be unnecessary.} which we set to $1$.
When the OOD signal is zero the model behaves like the baseline RNNLM, when it is high instead it behaves more like a unigram language model.
This way, we can retain the expressivity of "shortcut learning" when it is beneficial, and hopefully avoid its influence when it is detrimental.

\section{Experiments}
\label{SEC:EXPERIMENTS}


\paragraph{Setup}
\label{SEC:EXPERIMENTS:SETUP}

For all our language modelling experiments we use two-layer stacked GRUs, with a PyTorch implementation based on code by \citet{zhang-sennrich-2019-lightweight}\footnote{\url{https://github.com/bzhangGo/lrn}}.
We train separate models on the Penn Treebank and Wikitext-2 corpora using the default hyperparameters provided by the codebase.
We also train models on BPE subtokenized \citep{sennrich-etal-2016-neural} versions of the corpora using SentencePiece\footnote{\url{https://github.com/google/sentencepiece} 
}.
These models are used both as baselines and to provide the initial models for our approach.
For our approach we train one RND OOD model for each layer of the RNNLM, the teachers are 2-layer LeakyReLU MLPs \citep{Maas13rectifiernonlinearities} with layer normalization \citep{ba2016layer} and the students are like the teachers followed by 4 Resnet blocks \citep{He2016Resnet} with 2 LeakyReLU MLP layers each.
All hidden dimensions are set to match the RNNLM state dimension.
For consistency with the original codebase, we use SGD with decaying learning rate and early stopping to train the baseline RNNLMs, while we switch to Adam \citep{Kingma2015Adam}, with constant learning rate and early stopping when training the RND OOD estimator.
GRU hyperparameters are the default ones from the reported Penn Treebank and Wikitext-2 models of the baseline implementation.
The code to run the experiments is available.\footnote{\url{https://github.com/Avmb/lm-robustness}}

\paragraph{Perplexity estimation}
\label{SEC:EXPERIMENTS:PERPLEXITY}

\begin{table*}[ht!]
\centering
\small
\begin{tabular}{@{}llllllllll@{}}
\toprule
& in-domain & & \multicolumn{5}{c}{\cite{muller-etal-2020-domain}} & \multicolumn{2}{c}{MTNT}\\
\cmidrule(lr){2-2}
\cmidrule(lr){4-8}
\cmidrule(l){9-10}
& Penn & WT-2 & IT & Koran & Law & Med & Sub & fr-en.en & ja-en.en\\ 
\midrule
\multicolumn{10}{c}{word-level}\\
Baseline & 68.04 & \phantom{00}55.73 & \phantom{00}59.37 & \phantom{00}50.12 & \phantom{00}64.12 & \phantom{00}35.10 & \phantom{00}47.81 & \phantom{00}76.75 & \phantom{00}66.08\\
RND & \textbf{67.86}  & \phantom{00}\textbf{55.00} & \phantom{00}\textbf{58.18} & \phantom{00}\textbf{49.12} & \phantom{00}\textbf{62.94} & \phantom{00}\textbf{34.67} & \phantom{00}\textbf{47.01} & \phantom{00}\textbf{75.33} & \phantom{00}\textbf{64.73} \\

RND (abl.) & 67.84 & \phantom{00}55.41 & \phantom{00}59.02 & \phantom{00}49.76 & \phantom{00}63.67 & \phantom{00}34.99 & \phantom{00}47.55 & \phantom{00}76.25 & \phantom{00}65.64\\
\addlinespace
\hline
\addlinespace
\multicolumn{10}{c}{BPE-level}\\
Baseline& \textbf{27.85} & 1371.16 & 5657.39 & 5493.64 & 4123.78 & 5657.54 & 4048.14 & 2837.97 & 4051.66 \\
RND & 28.16 & \textbf{1197.07} & \textbf{4828.55} & \textbf{4774.78} & \textbf{3552.22} & \textbf{4520.38} & \textbf{3558.99} & \textbf{2519.75} & \textbf{3551.63}\\
RND (abl.) & 27.93 & 1287.26 & 5178.40 & 5159.36 & 3815.87 & 4898.91 & 3792.19 & 2675.51 & 3794.22 \\

\bottomrule

\end{tabular}

\caption{Perplexity of language models trained on the Penn Treebank dataset.}
\label{tab:rnd-results-penn-trained}
\end{table*}

\begin{table*}[ht!]
\centering
\small
\begin{tabular}{@{}llllllllll@{}}
\toprule
& in-domain & & \multicolumn{5}{c}{\cite{muller-etal-2020-domain}} & \multicolumn{2}{c}{MTNT}\\
\cmidrule(lr){2-2}
\cmidrule(lr){4-8}
\cmidrule(l){9-10}
& WT-2 & Penn & IT & Koran & Law & Med & Sub & fr-en.en & ja-en.en\\
\midrule
\multicolumn{10}{c}{word-level}\\
Baseline & \textbf{64.69} & 361.84 & 162.01 & 159.02 & 178.92 & 103.87 & \phantom{0}\textbf{96.65} & \textbf{177.73} & 184.69\\
RND & \textbf{64.69} & \textbf{333.52} & \textbf{156.73} & 156.59 & \textbf{171.42} & \textbf{102.33} & 100.46 & 175.34 & \textbf{180.54} \\
RND (abl.) & \textbf{64.69} & 338.33 & 157.96 & \textbf{155.75} & 172.94 & 102.74 & \phantom{0}98.82 & 174.20 & 180.91\\

\addlinespace
\hline
\addlinespace
\multicolumn{10}{c}{BPE-level}\\

Baseline & \textbf{29.39} & 190.91 & 648.84 & \textbf{694.86} & 339.46 & 355.74 & \textbf{563.92} & \textbf{495.39} & \textbf{497.27}\\
RND & 29.73 & \textbf{183.23} & 637.63 & 712.93 & 335.86 & 348.16 & 656.86 & 530.45 & 526.30\\
RND (abl.) & 29.46 & 185.48 & \textbf{632.52} & 695.54 & \textbf{334.37} & \textbf{347.75} & 624.27 & 515.16 & 512.96\\
\bottomrule

\end{tabular}

\caption{Perplexity of language models trained on the Wikitext-2 dataset.}
\label{tab:rnd-results-wt2-trained}
\end{table*}

\begin{table*}[ht!]
\centering
\small
\begin{tabular}{@{}llllllllll@{}}
\toprule
& & & \multicolumn{5}{c}{\cite{muller-etal-2020-domain}} & \multicolumn{2}{c}{MTNT}\\
\cmidrule(lr){2-3}
\cmidrule(lr){4-8}
\cmidrule(l){9-10}
Training & WT-2 & Penn & IT & Koran & Law & Med & Sub & fr-en.en & ja-en.en\\
\midrule
WT-2 & \textbf{0.0240}* & 0.1137 & 0.0735 & 0.1155 & 0.0767 & 0.0485 & 0.0936 & 0.1070 & 0.0896\\
\midrule
Penn & 0.0237 & 0.0252* & 0.0244 & \textbf{0.0233} & 0.0244 & 0.0234 & 0.0236 & 0.0240 & 0.0238\\
\midrule
WT-2 (BPE) & \textbf{0.0220}* & 0.0534 & 0.1054 & 0.1824 & 0.0697 & 0.0657 & 0.1472 & 0.1328 & 0.1196\\
\midrule
Penn (BPE) & \textbf{0.0256} & 0.0257* & 0.0321 & 0.0279 & 0.0313 & 0.0359 & 0.0300 & 0.0308 & 0.0302\\

\bottomrule
\end{tabular}
\caption{OOD estimates, averaged over GRU layers and tokens in each test set. * denotes the in-domain test sets.}
\label{tab:rnd-results-ood-scores}
\end{table*}

We investigate OOD performance with two standard corpora, Penn Treebank and Wikitext2. We evaluate each of the models both in-distribution, on the default test set of its training corpus, and out-of-distribution, on the test set of the other corpus. We also use  additional test sets adapted from machine translation robustness evaluations, specifically the English sides of the De-En test sets of \citet{muller-etal-2020-domain}, which is a collection of corpora from diffent domains (I.T., Koran, law, medical and movie subtitles) and the English sides of the MTNT Ja-En and Fr-En test sets of \citet{michel2018mtnt}, which are corpora scraped from Reddit and have been used for the WMT-19 robustness shared task \citep{li-EtAl:2019:WMT1}.

We report the results in tables \ref{tab:rnd-results-penn-trained} and \ref{tab:rnd-results-wt2-trained}.
We find that for the word-level models trained on Penn Treebank our approach improves the perplexity consistently both in-distribution and out-of-distribution for all the test sets we considered.
For the word-level models trained on Wikitext-2 our approach has a slightly worse perplexity in-distribution but improves on most OOD test sets, namely Penn Treebank test set, the Ja-En test set of the MTNT corpus and all the tests sets of \citet{muller-etal-2020-domain} except the Koran and subtitles test sets.
The Penn Treebank results are somewhat anomalous in that the perplexity of some OOD test sets is lower than the perplexity of the in-distribution test sets (and in fact the perplexity of the Wikitext-2 test set is even lower that the perplexity of the same test set evaluated by its own in-domain model).
This effect is caused by the limited vocabulary of the Penn Treebank training set which causes many of the tokens of the OOD test sets to be replaced by UNKs, which are easy to predict.
To avoid this artifact, we evaluate BPE-level models, which are open vocabulary and hence do not introduce any UNKs.
For the BPE-level models we find that for both the baselines and the RND approach the perplexities on the OOD datasets are much higher than the perplexities on the in-domain test sets.
Comparing our approach to the baselines, we observe a minimal degradation of perplexity in-distribution and substantial improvements on all OOD test sets when training on Penn Treebank, while when training on Wikitext-2 we observe more mixed results.

In order to analyse if the model is learning sensible values for scaling the out-of-distribution states, we compute the OOD scores estimated by the RND OOD detectors, averaged over the two GRU layers and over all the tokens in each test set.
We report these scores in table \ref{tab:rnd-results-ood-scores}.
The models trained on Wikitext-2 (both the word-level and BPE-level versions) always estimate the lowest OOD scores on the in-domain test set, as expected.
The Penn Treebank word-level model performs poorly, estimating similar scores for all the test sets, consistent with the aforementioned vocabulary collapse to UNKs, the BPE-level model instead is generally able to distinguish in-domain and out-of-domain test sets, albeit by a small margin and fails on one test set (Wikitext-2).

\paragraph{Ablation}
\label{SEC:EXPERIMENTS:ABLATION}

One could hypothesize that the improvements obtained by our model are due to just increasing the entropy of the output distribution rather than dropping unnecessary context from the RNN state.
We evaluate a variant of our model where we apply the OOD scaling only on the output of the top-layer RNN but not to the internal states. 
This increases the output entropy without affecting the context remembered by the model between time steps.
This ablation generally improves over the baseline but performs worse than our full model except for the model trained on Wikitext-2 BPE where the results are mixed.

\section{Conclusions and future work}
\label{SEC:CONCLUSIONS}

We proposed a method to improve the robustness of language models to distribution shift caused by train/test domain mismatch. 
Our model contracts the RNN state based on an unsupervised out-of-distribution estimator in order to reduce the model dependency on weak long-distance correlations, which are useful in-distribution but tend to be spurious in out-of-distribution conditions.
We obtain perplexity improvements on multiple out-of-domain test sets without substantial degradation on in-domain test sets.

While our approach is based on Recurrent decoders, its general principles may be applicable to other neural architectures.
For instance, the self-attention heads of a Transformer might modulated by an OOD detector in order to avoid attending to out-of-distribution parts of a sentence. 
We anticipate that extending our method to these kind of models will be a promising research direction.

\section*{Broader impact and ethical concerns}
\label{SEC:IMPACT}

This work provides improvements for language model technology on application domains not well represented in the training data. 

We expect that our approach might promote an increased deployment and usage of such technology. 
We do not expect our approach to introduce any bias against any specific group of users. 
Our approach adds only small computational costs over baseline language models and therefore is unlikely to prevent users with limited computational budgets from benefiting from the technology.

\section*{Acknowledgments}
\label{SEC:ACK}

This project received funding from the European Union’s Horizon 2020 research and innovation programme under grant agreement 825299 (GoURMET),
the European Research Council (ERC StG BroadSem 678254; ERC CoG TransModal 681760) and funding by the UK Engineering
and Physical Sciences Research Council (EPSRC)
fellowship grant EP/S001271/1 (MTStretch).


\bibliography{anthology,custom}

\clearpage
\appendix
\appendixpage
\section{Sequence-to-sequence experiments}
\label{SEC:SEQ2SEQ}
We performed additional experiments on sequence-to-sequence (seq2seq) tasks.
We obtained negative results, which we report here.

\paragraph{Architecture}
Our models use a Transformer-GRU architecture. 
The encoder is a standard bidirectional Transformer  while the decoder is a two-layer stacked GRU (sec. \ref{SEC:BACKGROUND:RNNLM}). The recurrent cell also accesses contextual embeddings of a source sentence tokens via an attention mechanism implemented as in \citet{luong-etal-2015-effective}, except that instead of a single attention head we use a Transformer multihead attention layer, similar to \citet{Chen2018GoogleTransl}.
The RND OOD model has the same architecture as in the LM experiments, although for simplicity we train it jointly with the MT models rather than in a separate stage, we make sure not to propagate gradients between the RND OOD model and the translation model hence there is no tradeoff between their training objectives.
The implementation is based on the Fairseq \citep{ott2019fairseq} Transformer and LSTM architectures, using the hyperparameters for their default IWSLT14 configuration.


\paragraph{Machine translation}
We trained De$\rightarrow$En translation models on the IWSLT14 training set \citep{cettolo2014report} with the standard Fairseq preprocessing pipeline\footnote{prepare-iwslt14.sh}.
We used on the standard test set produced by the preprocessing script as our in-domain test set and the \citet{muller-etal-2020-domain} test sets as our OOD test sets.
We report BLEU scores in table \ref{tab:mt-results}.
The baseline and the RND model have nearly identical scores on the in-domain test set, while they deviate up to about $1$ BLEU point on the OOD test sets, although in a non-systematic way.

\begin{table}[ht!]
\centering
\small
\begin{tabular}{@{}lllllll@{}}
\toprule
& in-domain & \multicolumn{5}{c}{\cite{muller-etal-2020-domain}} \\
\cmidrule(lr){2-2}
\cmidrule(lr){3-7}
& IWSLT14 & IT & Koran & Law & Med & Sub \\
\midrule
Base & 32.95 & 11.03 & \phantom{0}\textbf{5.72} & 11.35 & 13.76 & \textbf{19.19} \\
RND      & \textbf{32.97} & \textbf{12.07} & \phantom{0}5.13 & \textbf{12.02} & \textbf{13.93} & 18.71 \\
\bottomrule

\end{tabular}

\caption{Machine translation results}
\label{tab:mt-results}
\end{table}

\paragraph{Sentence reversal}
We considered a synthetic task intended to elicit the RND OOD activity.
The source segments consist each of a number of concatenated sentences separated by a separator token, the target segments are made of the same sentences, where each sentence is reversed at token level, but the sentences are concatenated in the same order as the source.
Since reversing a sentence does not depend on the previous sentences in the segment, the previous sentences become distractors that pollute the decoder GRU state with irrelevant information.
The model can learn to compensate in in-domain conditions where the test set is sampled from the same distribution of the training set, but we hypothesize that in OOD scenarios with longer segments composed by a higher number of sentences this spurious information will greatly decrease accuracy.
We test whether the RND OOD mechanism is effective at discarding this spurious information.

We consider two versions of the task, in one we sample the source segments from a synthetic vocabulary of $256$ tokens, with uniform probability per token, $32$ tokens per sentence, $8$ sentences per training segment.
We test in-domain at $8$ sentences and OOD at $10$ and $12$ sentences per segment.
In the second version, we train on concatenations of $4$ consecutive sentences of the English side of the IWSLT14 De-En training set, and we test at $4, 6, 8, 10$ and $12$ sentences per segment.
We use the same hyperparameters of our translation experiments, during inference we constrain the decoder to match the source length.

All the models achieve near perfect ($> 99.9$) BLEU scores in-domain, while OOD the scores quickly decrease as the number of sentences per segment increases, as expected.
Unfortunately we find no systematic difference between baseline and RND OOD models.

\paragraph{Discussion}
Unlike our language modeling experiments, we did not observe systematic improvements from using the RND out-of-distribution detector to contract the state of the GRU decoder in our sequence-to-sequence results.
There are multiple possible hypotheses for this discrepancy, such as encoder effects, generating outputs by beam search rather than scoring natural text, or the target distribution being more peaked around the mode.
We plan to investigate this effect in the future.

\end{document}